\documentclass[runningheads]{llncs}
\usepackage{graphicx}
\usepackage{amsmath}
\usepackage{todonotes}
\usepackage{siunitx}
\usepackage[export]{adjustbox}
\usepackage{booktabs}
\usepackage{amssymb}
\usepackage{enumitem}

\newcommand{\beginsupplement}{%
	\setcounter{table}{0}
	\renewcommand{\thetable}{S\arabic{table}}%
	\setcounter{figure}{0}
	\renewcommand{\thefigure}{S\arabic{figure}}%
}

\begin{document}
\title{Hierarchical brain parcellation with uncertainty}


\author{Mark S. Graham \inst{1} \and
Carole H. Sudre \inst{1} \and
Thomas Varsavsky \inst{1} \and
Petru-Daniel Tudosiu \inst{1} \and
Parashkev Nachev \inst{2} \and
Sebastien Ourselin \inst{1} \and
M. Jorge Cardoso \inst{1}}
\authorrunning{M.S. Graham et al.}
%
\institute{Biomedical Engineering and Imaging Sciences, King’s College London, UK \and
Institute of Neurology, University College London, London, United Kingdom  \\
\email{mark.graham@kcl.ac.uk}}

\maketitle              
\begin{abstract}
Many atlases used for brain parcellation are hierarchically organised, progressively dividing the brain into smaller sub-regions. However, state-of-the-art parcellation methods tend to ignore this structure and treat labels as if they are `flat'. We introduce a hierarchically-aware brain parcellation method that works by predicting the decisions at each branch in the label tree. We further show how this method can be used to model uncertainty separately for every branch in this label tree. Our method exceeds the performance of flat uncertainty methods, whilst also providing decomposed uncertainty estimates that enable us to obtain self-consistent parcellations and uncertainty maps at any level of the label hierarchy. We demonstrate a simple way these decision-specific uncertainty maps may be used to provided uncertainty-thresholded tissue maps at any level of the label tree.
\end{abstract}

\section{Introduction}
Brain parcellation seeks to partition the brain into spatially homogeneous structural and functional regions, a task fundamental for allowing us to study the brain in both function and dysfunction. The brain is hierarchically organised, with smaller subregions performing increasingly specialised functions, and the atlases classically used for parcellation typically reflect this by defining labels in a hierarchical tree structure. Manual parcellation is also typically performed hierarchically; typically semi-automated methods are used to help delineate larger structures with sufficient tissue contrast, and these are then manually sub-parcellated using anatomical or functional landmarks \cite{neuromorphometricsProtocol}.  

The state-of-the-art for brain parcellation has come to be dominated by convolutional neural networks (CNNs). These methods tend to ignore the label hierarchy, instead adopting a `flat' label structure. However, methods that are aware of the label hierarchy are desirable for many reasons. Such methods could degrade their predictions gracefully, for example labelling a noisy region with the coarser label `cortex' rather then trying to assign a particular cortical division. They also offer the opportunity to train on multiple datasets with differing degrees of label granularity, assuming those labels can be mapped onto a single hierarchy.
 
Hierarchical methods also enable uncertainty to be modelled at different levels of the label tree. There has been recent interest in using uncertainty estimates provided by CNNs \cite{kendall2017uncertainties,kendall2018multi} to obtain confidence intervals for downstream biomarkers such as regional volumes \cite{eaton2018towards,wang2019automatic}, which is key if these biomarkers are to be integrated into clinical pipelines. Flat methods provide only a single uncertainty measure per voxel, which prevents attribution of the uncertainty to a specific decision. Hierarchical methods can provide uncertainty for each decision along the label hierarchy, for example enabling the network to distinguish between relatively easy decisions (e.g. cortex vs non-cortex) and more challenging decisions, such as delineating cortical sub-regions that are ill-defined on MRI. This could facilitate more specific and informative confidence bounds for derived biomarkers used in clinical decision making.


 
 
Whilst hierarchical methods have been applied to classification, \cite{demyanov2017tree,redmon2017yolo9000,hu2016learning,wu2019hierarchical}, there are very few CNN-based methods that attempt hierarchical segmentation. A method proposed by Liang et al. \cite{liang2018dynamic} has been applied to perform hierarchical parcellation of the cerebellum \cite{han2019hierarchical}. A drawback of this approach is that the  tree structure is directly built into the model architecture, requiring a tailored model to be built for each new label tree. 

In this work we make two contributions. Firstly, we extend a method previously proposed for hierarchical classification \cite{redmon2017yolo9000} to hierarchically-aware segmentation. The method works by predicting decisions at each branch in the label tree, and has the advantage that it requires no alteration to the network architecture. Secondly, we show it is possible to use such a model to estimate uncertainty at each branch in the label tree. Our model with uncertainty matches the performance of `flat' uncertainty methods, whilst providing us with decomposed uncertainty estimates that enable us to obtain consistent parcellations with corresponding uncertainty at any level of the label tree. We demonstrate how these decision-specific uncertainty maps can be used to provide uncertainty-thresholded tissue segmentations at any level of the label tree.

\section{Methods}

\begin{figure}[!b]
	\centering
	\includegraphics[width=1\textwidth,trim={0 0 0 0},clip,center]{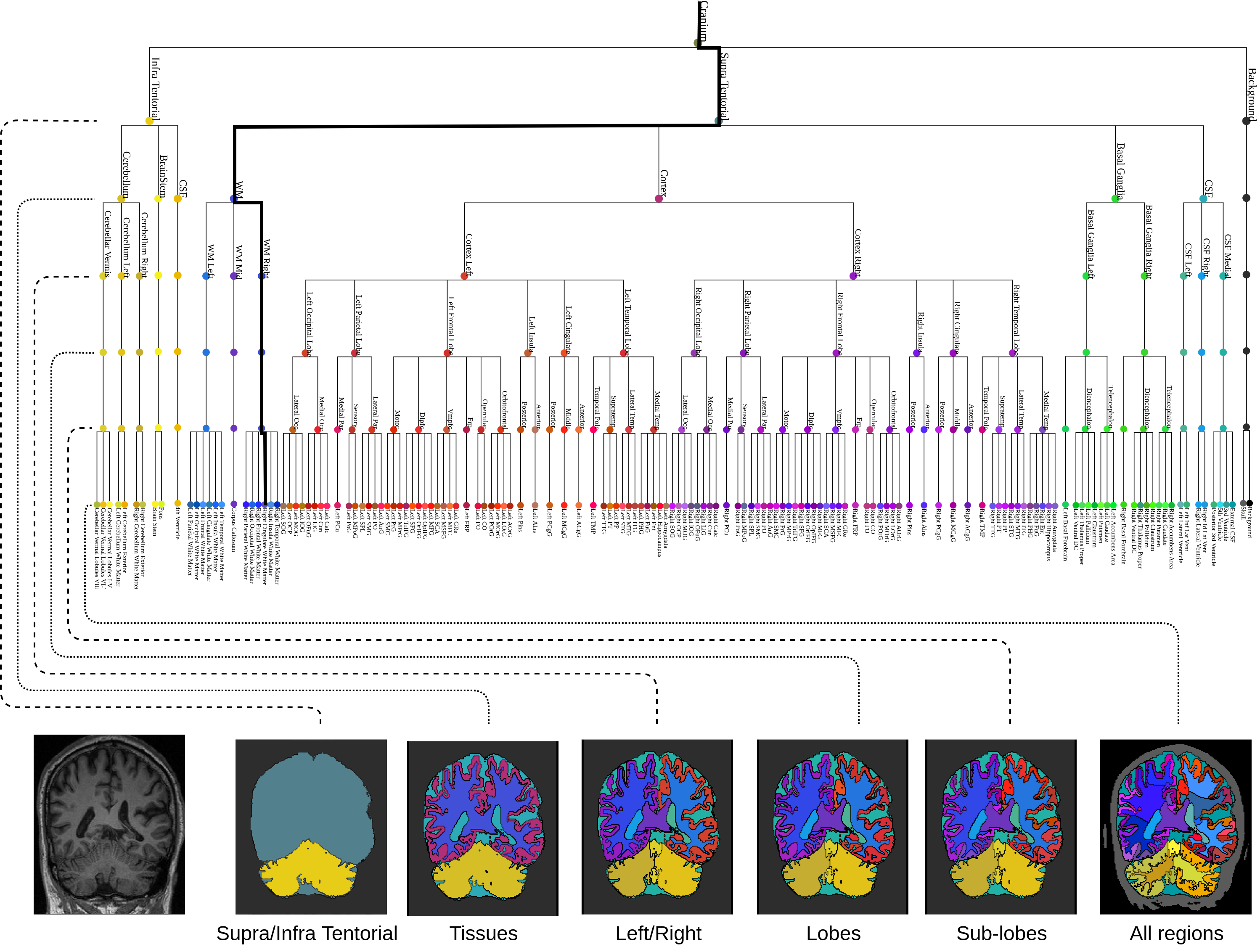}
	\caption{The neuro-anatomical label hierarchy considered in this paper, with the path from the root to the right cingulate highlighted.  A larger version of this tree is included in the supplementary materials.}
	\label{fig:figure1_tree}
\end{figure}

We first review existing flat segmentation models with uncertainty, before describing how we apply an existing classification model to perform hierarchical parcellation. We then show how such a model can be used to provide hierarchical uncertainty estimates. We focus on modelling intrinsic uncertainty in this work, although the methods presented can be straightforwardly extended to estimating model uncertainty, too.

\subsection{Flat parcellation}
In a flat segmentation scenario, we consider the task as per-voxel classification, where the likelihood for a voxel is given by $p(\mathbf{y}|\mathbf{W},\mathbf{x}) = \mathrm{Softmax}\left(\mathbf{f}^{\mathbf{W}}(\mathbf{x})\right)$ where $\mathbf{f}^{\mathbf{W}}(\mathbf{x})$ is the output of a neural network with weights $\mathbf{W}$, input $\mathbf{x}$ is a 3D image volume, and $\mathbf{y}$ encodes the $C$ segmentation classes. We seek the weights $\mathbf{W}$ that minimise the negative log-likelihood, yielding the standard cross-entropy loss function, $ \mathrm{CE}\left(y=c, \mathbf{f}^{\mathbf{W}}(\mathbf{x})\right) = -  \log \mathrm{Softmax}\left(f_c^{\mathbf{W}}(\mathbf{x})\right)$.
As in Kendall et al. \cite{kendall2017uncertainties}, heteroscedastic intrinsic uncertainty can be modelled by considering scaling the logits by a second network output, $\sigma^{\mathbf{W}}(\mathbf{x})$, giving a likelihood of $p(\mathbf{y}|\mathbf{W},\mathbf{x}, \sigma) = \mathrm{Softmax}\left(\frac{1}{\sigma^2(\mathbf{x})}\mathbf{f}^{\mathbf{W}}(\mathbf{x})\right)$. $\sigma^{\mathbf{W}}(\mathbf{x})$ is a per-voxel estimate, so it has the same dimension as $\mathbf{x}$. Employing the approximation
$\frac{1}{\sigma^{\mathbf{W}}(\mathbf{x})^2}\sum_c \exp{\left(\frac{1}{\sigma^{\mathbf{W}}(\mathbf{x})^2}f_c^{\mathbf{W}}(\mathbf{x})\right)} 
\approx  \left( \sum_c \exp{\left(f_c^\mathbf{W}(\mathbf{x})\right)}\right)^{\sigma^{\mathbf{W}}(\mathbf{x})^{-2}}$ used in \cite{kendall2018multi} allows us to write the negative log-likelihood as 

$$ \mathcal{L}(y=c,\mathbf{x}; \mathbf{W}) = \frac{\mathrm{CE}\left(y=c, \mathbf{f}^{\mathbf{W}}(\mathbf{x})\right)}{\sigma^{\mathbf{W}}(\mathbf{x})^2} + \log{ \sigma^{\mathbf{W}}(\mathbf{x})} $$

\subsection{Hierarchical parcellation}
Here we describe the hierarchical classification/detection model proposed by Redmon et al. \cite{redmon2017yolo9000}, and discuss how it can be adapted for segmentation tasks. The methods described here are general to all label taxonomy trees, but in this work we specifically consider the tree shown in Figure~\ref{fig:figure1_tree}, described in more detail in Section~\ref{section:data}. The probabilities at each node obey simple rules: the probabilities of all a node's children sum to the probability of the node itself, and so if we take $p(\text{root})=1$ the probabilities of all leaf nodes sum to 1. Leaf node probabilities can be expressed as the product of conditional probabilities down the tree; for example using the hierarchy in Figure~\ref{fig:figure1_tree} we can express $p(\textrm{Right cingulate WM})$ as
\begin{flalign*}
  \begin{aligned}
p(\textrm{Right cingulate WM}) = &p(\textrm{Right cingulate}|\textrm{Right WM})p(\textrm{Right WM}|\textrm{WM})\ldots \\
&p(\textrm{WM}|\textrm{Supra tentorial})p(\textrm{Supra tentorial}|\textrm{Cranium})\ldots\\
&p(\textrm{Cranium})
 \end{aligned}
\end{flalign*}
 
\noindent where $p(\textrm{Cranium})=1$. Our model predicts the conditional probabilities for each node, and is optimised using a cross-entropy loss at every level of the tree.


More formally, we label each node $i$ at level $l$ as $N_{i,l}$, where $l=0$ denotes the root and $l=L$ the deepest level, giving a maximum height of $L+1$.  Our model $\mathbf{f}^{\mathbf{W}}(\mathbf{x})$ produces a score for each node in the tree, $f^{\mathbf{W}}(\mathbf{x})_{i,l}$. We define a hierarchical softmax - essentially a softmax over the siblings for a given node - to produce the conditional probabilities at each node,

$$ p_{i,l} = \dfrac{\exp\left( f^{\mathbf{W}}(\mathbf{x})_{i,l}\right)}{\sum_{N_{j,l}=S[N_{i,l}]}\exp\left( f^{\mathbf{W}}(\mathbf{x})_{j,l}\right)}$$
where $S[N_{i,l}]$ denotes all the sibling nodes of $N_{i,l}$, including itself.

In the flat case we had a single label per voxel, $y_c$. In the hierarchical case $y_c$ denotes a leaf node of the tree, and we consider the label superset $A[y_c] = \{N_{i,l}\}$ comprising all the nodes traversed from the root to the label's leaf node, excluding the root node but including itself. The total loss is the summation of a CE loss calculated at each level of the tree,

$$\mathcal{L}\left(y=c,\mathbf{x}; \mathbf{W}\right) = -\sum_{N_{i,l} \in A[y_c]}  \log p_{i,l}$$

For parcellation the network makes a prediction per voxel, that is $\mathbf{f}^{\mathbf{W}}(\mathbf{x}) \in \mathbb{R}^{x \times y \times z \times H}$ where $H$ is the total number of nodes, making the considerably more computationally expensive than in classification tasks. The denominator of the hierarchical softmax can be efficiently calculated as a matrix multiplication, allowing $p_{i,l}$ to be calculated from the elementwise division of two matrices.

\subsection{Hierarchical uncertainty}
We extend the model by modelling an uncertainty for every decision made along the tree. The network output $\boldsymbol{\sigma}^{\mathbf{W}}(\mathbf{x})$ is now vector-valued, and exists for every non-leaf node, $\sigma^{\mathbf{W}}(\mathbf{x})_{i,l}$. The loss becomes:

$$\mathcal{L}\left(y=c,\mathbf{x}; \mathbf{W}\right) = -\sum_{N_{i,l} \in A[y_c]}  \dfrac{ \log p_{i,l}}{\sigma^{\mathbf{W}}(\mathbf{x})_{i,l-1}^2} + \log \sigma^{\mathbf{W}}(\mathbf{x})_{i,l-1}$$

\noindent In this formulation the uncertainty values in a given voxel are unconstrained if they do not fall along the decision path for that voxel; for example values of $\sigma$ relating to cortical parcellation do not enter into the loss in white matter voxels. We add a penalty term to encourage shrinking every value of $\sigma_{i,l}$ that does not fall along the path from the true leaf node to the root node, giving a final loss of

\begin{flalign}
  \begin{aligned}
  \mathcal{L}\left(\mathbf{y}=c,\mathbf{x}; \mathbf{W}\right) 
=& -\sum_{N_{i,l} \in A[y_c]}  \left( \dfrac{ \log p_{i,l}}{\sigma^{\mathbf{W}}(\mathbf{x})_{i,l-1}^2} 
+ \log \sigma^{\mathbf{W}}(\mathbf{x})_{i,l-1} \right) \\
&+ \lambda \sum_{N_{i,l} \notin A[y_c]} \log \sigma^{\mathbf{W}}(\mathbf{x})_{i,l-1}
 \end{aligned}
\end{flalign}
where $\lambda$ controls the strength of this penalty.

\subsection{Architecture and implementation details}
The network is a 3D UNet based on the implementation described in the nnUNet paper \cite{nnUnet} and implemented in PyTorch. Our implementation contains three pooling layers and separate, identical decoder branches for the segmentation and uncertainty outputs. The parcellation branch predicts an output for each leaf node in the tree for the flat case - 151 for the tree considered in this work - and in the hierarchical case predicts an output for each node in the tree. As the hierarchical network does not make any predictions for nodes with no siblings, as $p(\text{node}|\text{parent})$=1 always for such nodes, the hierarchical model predicts 213 outputs per voxel for the same tree. The uncertainty branch predicts a single channel for flat models, and a number of channels equal to the number of branches in the label tree for hierarchical models - 61 for the tree in this work.  In practice, $\log(\sigma^2)$ is predicted for numerical stability. We set the penalty term in the hierarchical loss $\lambda=0.1$. Networks were trained on $110^3$ patches randomly sampled from the training volume. Group normalisation was used, enabling a batch size of 1 to be coupled with gradient accumulation to produce an effective batch size of 3. Models were trained with the Adam optimiser \cite{kingma2014adam} using a learning rate of $4e^{-3}$. Each model was trained for a maximum of 300 epochs with early stopping if the minimum validation loss did not improve for 15 epochs.

\section{Experiments and Results}
\subsection{Data}
\label{section:data}
We use the hierarchical label tree from the GIF label-fusion framework \cite{cardoso2015geodesic}, which is based on the labelling from the MICCAI 2012 Grand Challenge on label fusion \cite{landman2012miccai}. In total, there are 151 leaf classes and a hierarchical depth of 6, see Figure~\ref{fig:figure1_tree}.

We use 593 T1-weighted MRI scans from the ADNI2 dataset \cite{petersen2010alzheimer}, with an average voxel size of 1.18$\times$1.05$\times$1.05\si{mm\cubed} and dimension 182$\times$244$\times$246. Images were bias-field corrected, oriented to a standard RAS orientation and cropped using a tight mask. Silver-standard labels were produced using GIF on multimodal input data, followed by manual quality control and editing where necessary. 543 scans were used for training and validation, and 50 were reserved for testing.

\subsection{Experiments}
We consider the following  four models: 1) a baseline network trained on flat labels with weighted cross-entropy ($F$) 2) the same as ($F$) but with uncertainty estimates ($F_{\text{unc}}$), 3) a network trained on hierarchical labels ($H$), 4) a hierarchically-trained network with hierarchical uncertainty estimates ($H_{\text{unc}}$). The following experiments were performed: 
\begin{itemize}[noitemsep,topsep=0pt]
	\item Performance comparison using dice overlap on the withheld test data at all six levels of the tree.
	\item Qualitative assessment of the uncertainty maps provided by $H_{\text{unc}}$ and $F_{\text{unc}}$.
	\item Comparison of uncertainty-thresholded segmentations from $H_{\text{unc}}$ and $F_{\text{unc}}$.
\end{itemize}

\subsection{Results \& Discussion}
Dice scores for all the models are reported in Table~\ref{table1}. Despite predicting a tree-structure with $>41\%$ more predictions per voxel than the flat model, performance for $H$ only drops marginally when compared to $F$, consistent with existing performance comparisons between flat and hierarchical models in classification and object detection settings \cite{redmon2017yolo9000}. $H_{\text{unc}}$ outperforms $F_{\text{unc}}$ for the four more fine-grained levels of the label tree. This is likely due to the empirically observed difficult in stably training $F_{\text{unc}}$; we found no such problems with $H_{\text{unc}}$, which was easy to optimise.
\begin{table}[t!]
\centering   
\caption{Dice scores averaged over all classes on the test set for the flat ($F$) and the proposed hierarchical ($H$) model. Uncertainty-aware models are denoted with an $\text{unc}$ subscript. Values are Median (IQR) across the 50 subjects in the test set. Bold indicates significantly better performance between model pairs ($F$ vs $H$, $F_{\text{unc}}$ vs $H_{\text{unc}}$), at p$<0.05$, using p-values obtained from a Wilcoxon paired test.}    
\begin{adjustbox}{center}
\begin{tabular}{lccc|ccc}     
\toprule           
Tree level & $F$ & $H$ (ours) & $P$ & $F_{\text{unc}}$  & $H_{\text{unc}}$ (ours) & $P$ \\    
\midrule     
Supra/Infra & \textbf{0.986 (0.003)} & 0.985 (0.002) & $<$0.00005 & \textbf{0.984 (0.003)} & 0.984 (0.002) & 0.009 \\     
Tissue & \textbf{0.942 (0.007)} & 0.941 (0.008) & $<$0.00005 & 0.934 (0.007) & 0.934 (0.007) & 0.95 \\ 
Left/right & \textbf{0.942 (0.006)} & 0.938 (0.008) & $<$0.00005 & 0.932 (0.005) & \textbf{0.933 (0.006)} & 0.006 \\    
Lobes   & \textbf{0.924 (0.008)} & 0.922 (0.009) & 0.00001 & 0.913 (0.008) & \textbf{0.917 (0.008)} & $<$0.00005 \\     
Sub-lobes  & \textbf{0.891 (0.011)} & 0.884 (0.013) & $<$0.00005 & 0.870 (0.013) & \textbf{0.880 (0.012)} & $<$0.00005 \\     
All regions & \textbf{0.861 (0.011)} & 0.848 (0.015) & $<$0.00005 & 0.831 (0.018) & \textbf{0.845 (0.013)} & $<$0.00005 \\     \bottomrule     \end{tabular}
\end{adjustbox}
\label{table1}
\end{table}

\begin{figure}
	\centering
	\includegraphics[width=1\textwidth,trim={0 0 0 0},clip,center]{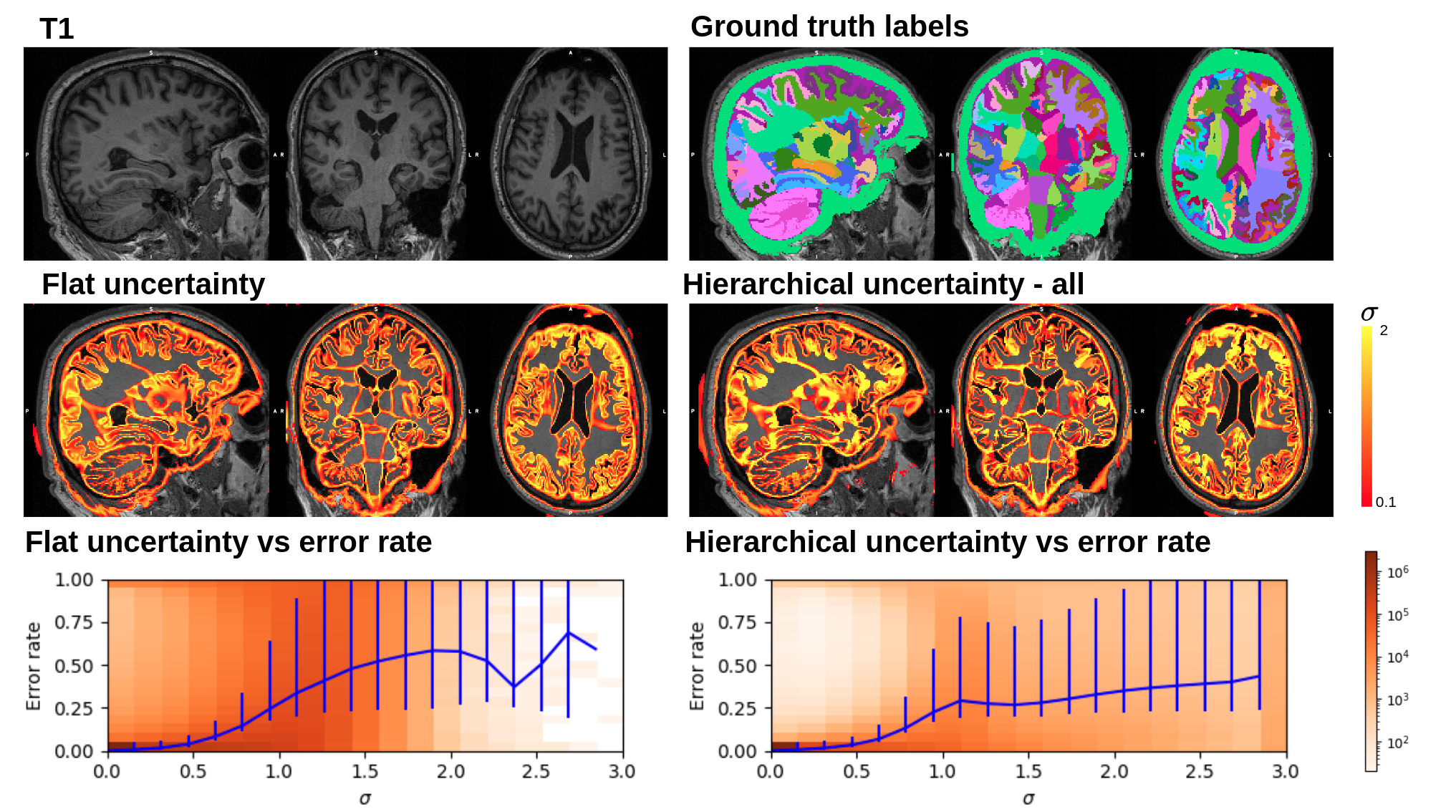}
	\caption{Evaluation of the uncertainty from $F_{\text{unc}}$ and the total uncertainty for $H_{\text{unc}}$ obtained by summing all uncertainty components. Joint histograms show voxel counts for $\sigma$ against (1-predicted probability for true class), averaged across all test subjects. Blue lines represent the mean error rate and error bars are 25-75 percentiles.}
	\label{fig:figure2}
\end{figure}

Figure~\ref{fig:figure2} compares the uncertainty map from $F_{\text{unc}}$ with the total uncertainty map from $H_{\text{unc}}$, obtained by summing all uncertainty components at each voxel. They look visually similar, and the joint histograms demonstrate expected trade-offs between uncertainty and error rate. Ideally, we would see low counts in the top-left of the joint histograms, indicating the models do not make confidently wrong predictions with low uncertainty. We see this desired behavior for $H_{\text{unc}}$ more strongly than $F_{\text{unc}}$.

Figure~\ref{fig:figure3} shows uncertainty maps predicted by $H_{\text{unc}}$ for different branches of the label tree. The model provides sensibly decomposed uncertainty maps for each decision along the label tree, with uncertainty strongly localised along decision boundaries. The maps reflect the uncertainty we expect for different decisions: for example there is highly localised uncertainty along the well contrasted WM-CSF boundary, but uncertainty is more spread out on boundaries between cortical regions which are poorly defined, and subject to high inter-rater variability.

\begin{figure}
	\centering
	\includegraphics[width=1\textwidth,trim={0 0 0 0},clip,center] {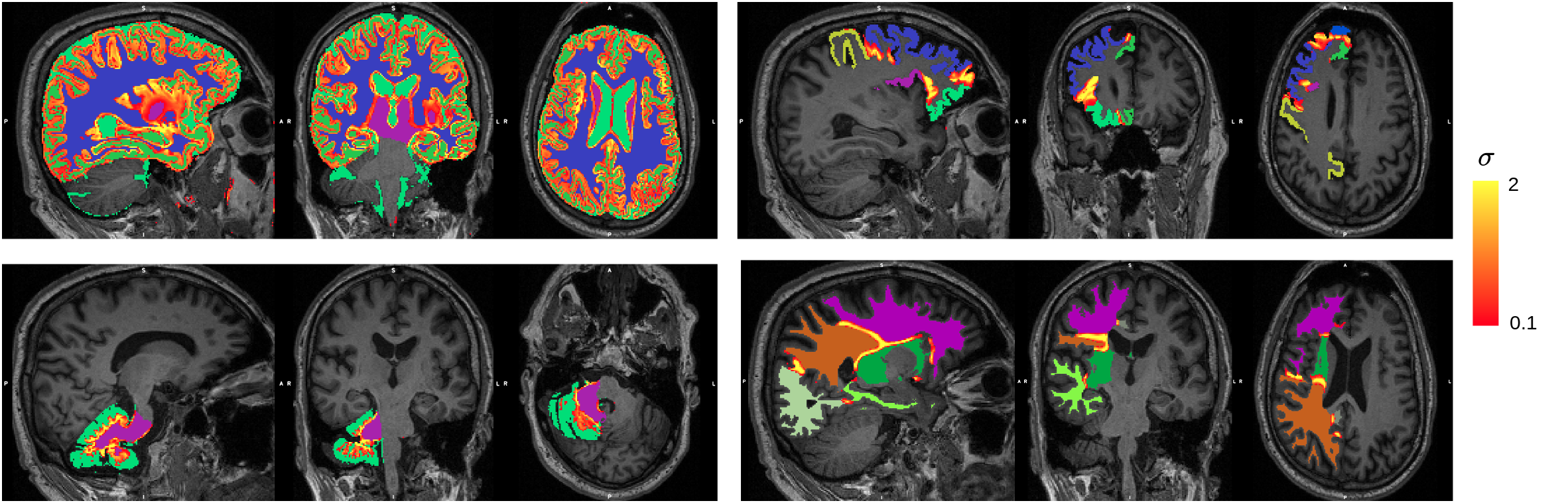}
	\caption{Demonstration of different uncertainty components for model $H_{\text{unc}}$ at four different branches of the label hierarchy, shown alongside the tissue class options at that branch.  Colours have been selected to maximise distinguishability between adjacent classes.}
	\label{fig:figure3}
\end{figure}

\begin{figure}
	\centering
	\includegraphics[width=1\textwidth,trim={0 0 0 0},clip,center] {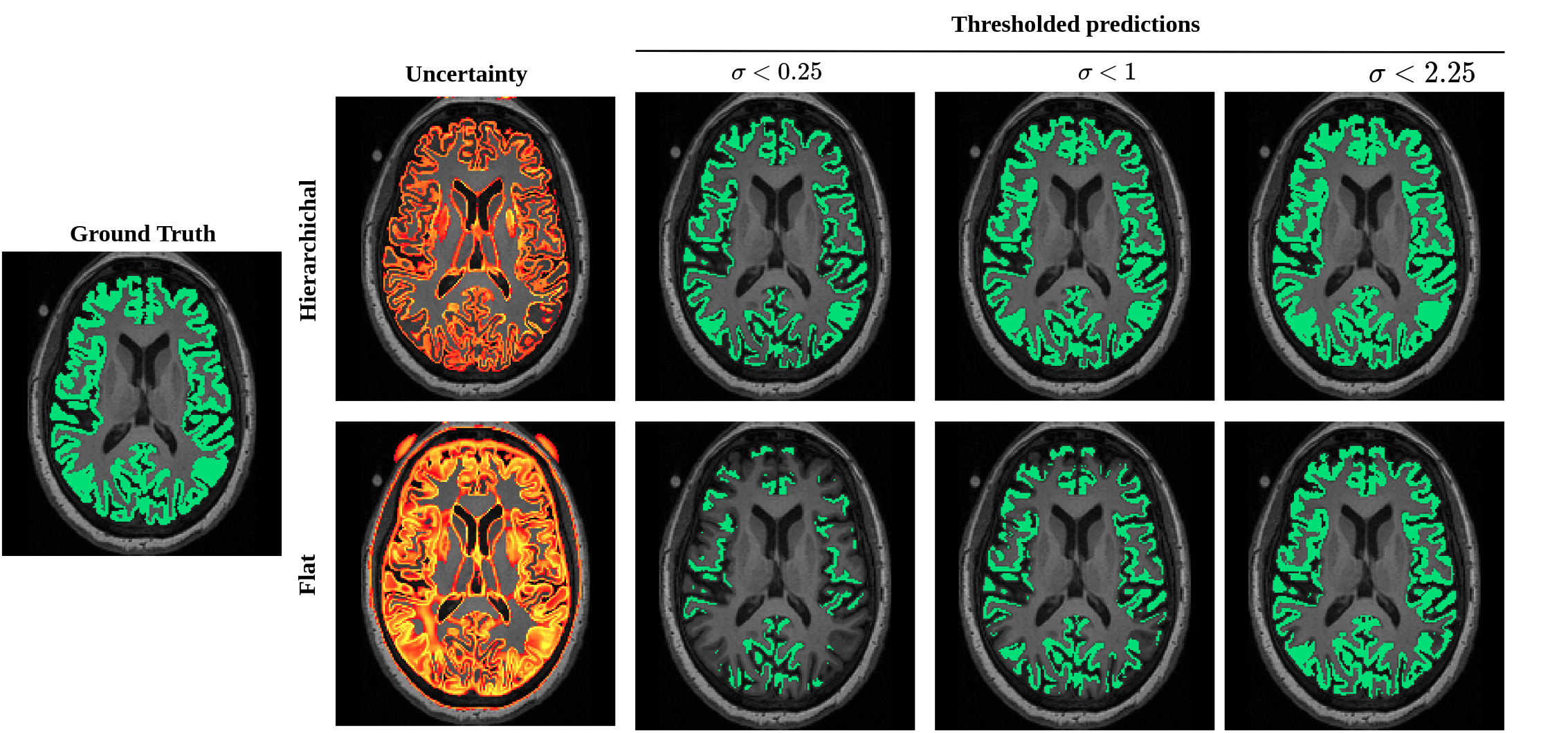}
	\caption{Demonstration of thresholding predictions according to uncertainty. Ground truth cortical segmentation is shown on left. Using $H_{\text{unc}}$  a cortex-specific uncertainty map can be produced,  that can be sensibly thresholded to create cortical predictions at different uncertainty levels. The lack of decision specificity in the single uncertainty map provided by $F_{\text{unc}}$ means we cannot perform cortex-specific thresholding - see in particular the map thresholded at $\sigma<0.25$.}
	\label{fig:figure4}
\end{figure}

Figure~\ref{fig:figure4} demonstrates a simple uncertainty-based thresholding method to obtain  upper- and lower-bound cortical maps. They show that the cortical-specific uncertainty component from $H_{\text{unc}}$ can be used to sensibly threshold predictions for non-leaf classes, in a way that is not possible for the uncertainty map from $F_{\text{unc}}$ which lacks specificity to non-leaf nodes.

\section{Conclusions}
We have proposed a hierarchically-aware parcellation model, and demonstrated how it may be used to produce per-decision measures of uncertainty on the label tree. Our method outperforms the flat uncertainty model in terms of dice score, and was less likely than the flat model to make wrong predictions with both high confidence and low uncertainty. Furthermore we demonstrate the decomposed uncertainty enables us to produce consistent parcellations along with uncertainty maps for classes higher up the label tree, which is not possible with flat uncertainty models.

%
%
\bibliographystyle{splncs04}
\bibliography{bibliography}

\beginsupplement
\begin{figure}[!ht]
	\centering
	\includegraphics[width=0.75\textwidth,trim={0 0 0 0},clip,center] {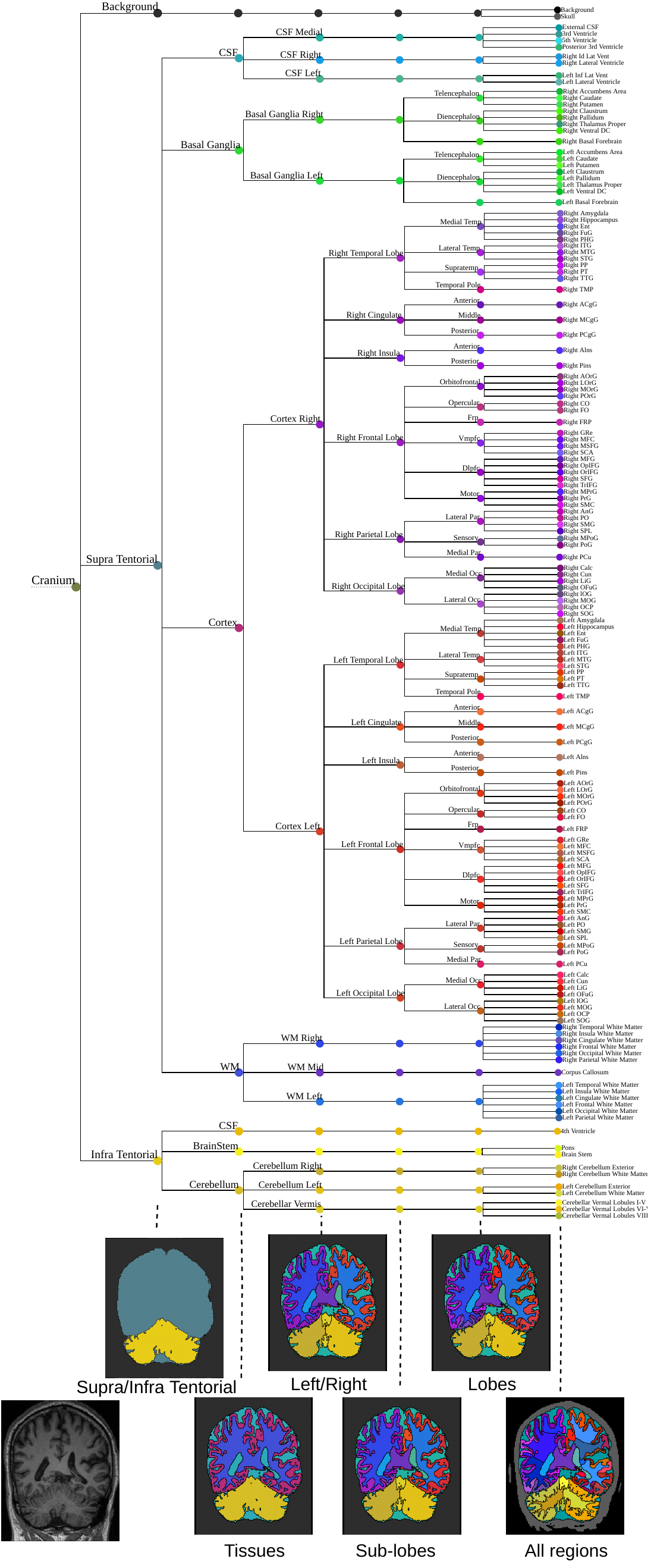}
	\caption{Larger version of the label hierarchy considered in this paper.}
	\label{fig:figure3}
\end{figure}
\end{document}